\documentclass{article}

\usepackage[final]{corl_2024} %

\usepackage{multicol}
\usepackage{graphics} %
\usepackage{epsfig} %
\usepackage{times} %
\usepackage{amsmath}
\usepackage{amssymb,bm}
\usepackage{stfloats}
\usepackage{subcaption}
\usepackage{url}
\usepackage[utf8]{inputenc}
\usepackage{algorithm}
\usepackage[noend]{algpseudocode}
\usepackage{dsfont}
\usepackage{enumitem}
\usepackage{wrapfig}

\hypersetup{
    colorlinks=true,
    linkcolor=blue,
    filecolor=magenta,      
    urlcolor=cyan,
    pdftitle={Overleaf Example},
    pdfpagemode=FullScreen,
    }

\title{D$^3$Fields: Dynamic 3D Descriptor Fields for Zero-Shot Generalizable Rearrangement}

\author{\textbf{Yixuan Wang$^{1*}$, Mingtong Zhang$^{2*}$, Zhuoran Li$^{3*}$, Tarik Kelestemur$^4$,}\\
\textbf{Katherine Driggs-Campbell$^2$, Jiajun Wu$^5$, Li Fei-Fei$^5$, Yunzhu Li$^1$}\\
\\
$^1$Columbia University, $^2$University of Illinois, Urbana-Champaign, $^3$National University of Singapore,\\
$^4$Boston Dynamics AI Institute, $^5$Stanford University}

\begin{document}

\maketitle

\newcommand{\change}[1]{{\textcolor{blue}{#1}}}
\newcommand{\rebuttal}[1]{{\textcolor{blue}{#1}}}
\newcommand{\yx}[1]{{\textcolor{blue}{[Yixuan: #1]}}}
\newcommand{\todo}{\textcolor{purple}{\textbf{TODO:}}}
\newcommand{\mt}{\textcolor{blue}{\textbf{Mingtong:}}}

\vspace{-25pt}

\begin{abstract}
Scene representation is a crucial design choice in robotic manipulation systems. An ideal representation is expected to be 3D, dynamic, and semantic to meet the demands of diverse manipulation tasks. However, previous works often lack all three properties simultaneously. In this work, we introduce D$^3$Fields---\textbf{dynamic 3D descriptor fields}.
These fields are \textbf{implicit 3D representations} that take in 3D points and output semantic features and instance masks.
They can also capture the dynamics of the underlying 3D environments.
Specifically, we project arbitrary 3D points in the workspace onto multi-view 2D visual observations and interpolate features derived from visual foundational models. The resulting fused descriptor fields allow for flexible goal specifications using 2D images with varied contexts, styles, and instances.
To evaluate the effectiveness of these descriptor fields, we apply our representation to rearrangement tasks in a zero-shot manner.
Through extensive evaluation in real worlds and simulations, we demonstrate that D$^3$Fields are effective for \textbf{zero-shot generalizable} rearrangement tasks.
We also compare D$^3$Fields with state-of-the-art implicit 3D representations and show significant improvements in effectiveness and efficiency.
\href{https://robopil.github.io/d3fields/}{Project Page}

\end{abstract}

\vspace{-15pt}

\begin{figure*}[!ht]
    \centering
    \vspace{-15pt}
    \includegraphics[width=\linewidth]{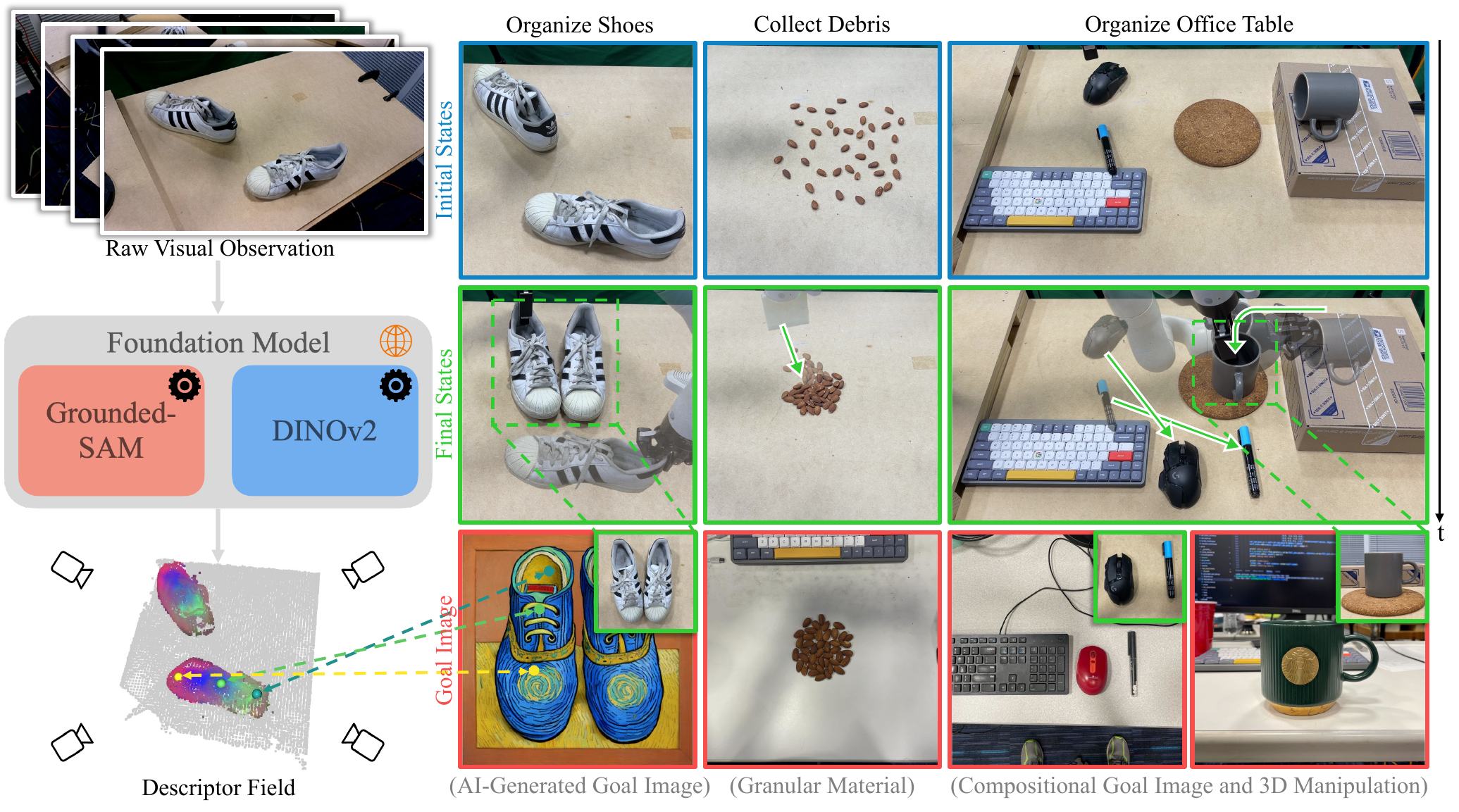}
    \vspace{-15pt}
    \captionof{figure}{\small
    \textbf{D$^3$Fields Representation and Application to Zero-Shot Rearrangement Tasks.}
    D$^3$Fields take in multi-view RGBD images and encode semantic features and instance masks using foundational models.
    The descriptor fields visualized in the bottom left using Principal Component Analysis (PCA) demonstrate consistent features across instances.
    We use our representation for rearrangement tasks given 2D goal images with diverse instances and styles in a zero-shot manner.
    We address pick-and-place tasks such as shoe organization and tasks requiring dynamic modeling like collecting debris.
    We also show that our framework can accomplish 3D manipulation and compositional task specification in the table organization task.
    }
    \vspace{-20pt}
    \label{fig:teaser}
\end{figure*}

\section{Introduction}

\vspace{-5pt}

The choice of scene representation is essential in robotic systems.
An ideal representation is expected to be simultaneously 3D, dynamic, and semantic to meet the needs of various robotic manipulation tasks in our daily lives.
However, previous research on scene representations in robotics often does not encompass all three properties. Some representations exist in 3D space~\cite{manuelli2020keypoints,li20213d,shi2022robocraft, Ze2022rl3d}, yet they overlook semantic information. Others focus on dynamic modeling~\cite{hafner2019learning,pmlr-v155-yan21a, Wang-RSS-23, minderer2019unsupervised}, but only consider 2D data, neglecting the role of 3D space. Some other works are limited by only considering semantic information such as object instance and category~\cite{tremblay2018deep,tyree20226,Jatavallabhula-RSS-23,mazur2023feature, structdiffusion2023}.

In this work, we aim to satisfy all three criteria by introducing D$^3$Fields, unified descriptor fields that are \textbf{3D, dynamic, and semantic}.
Notably, D$^3$Fields are \textbf{implicit 3D representations} rather than explicit 3D representations like point clouds.
D$^3$Fields take arbitrary 3D coordinates as inputs and output both geometric and semantic information corresponding to these positions. This includes the instance mask, dense semantic features, and the signed distance to the object surface.
Notably, deriving these descriptor fields requires no training and is conducted in a zero-shot manner, utilizing large visual foundation models and vision-language models (VLMs).
In our approach, we employ a set of advanced models. We first use Grounding-DINO~\cite{liu2023grounding}, Segment Anything (SAM)~\cite{kirillov2023segany}, XMem~\cite{cheng2022xmem}, and DINOv2~\cite{oquab2023dinov2} to extract information from multi-view 2D RGB images.
We then project arbitrary 3D coordinates back to each camera, interpolate to compute representations from each view, and fuse these data to derive the descriptors associated with these 3D positions, as shown in Figure~\ref{fig:teaser} (left).
Leveraging the dense semantic features and instance mask of our representation, we achieve robust tracking 3D points of the target object instances and train the dynamics models. These learned dynamics models can be incorporated into a Model-Predictive Control (MPC) framework to plan for zero-shot generalizable rearrangement tasks.

Notably, the derived representations allow for zero-shot generalizable rearrangement tasks, where the goal is specified by 2D images sourced from the Internet, smartphones, or even generated by AI models.
Such goal images have been challenging to manage with previous methods, because they contain varied styles, contexts, and object instances different from the robot's workspace.
Our proposed D$^3$Fields can establish dense correspondences between the robot workspace and the target configurations.
Given correspondences, we can define our planning cost and use the MPC framework with the learned dynamics model to derive actions for accomplishing tasks.
Remarkably, this task execution process does not require any further training, offering a highly flexible and convenient interface for humans to specify tasks for the robots.

We evaluate our method across a wide range of robotic rearrangement tasks in a zero-shot manner. These tasks include organizing shoes, collecting debris, and organizing office desks, as shown in Figure~\ref{fig:teaser} (right).
Furthermore, we provide both quantitative and qualitative comparisons with state-of-the-art implicit 3D representations to demonstrate the effectiveness and efficiency of our approach~\cite{ye2023featurenerf, shen2023distilled}.
Through a detailed analysis of our D$^3$Fields, we offer insights into the category-level generalization capabilities and zero-shot rearrangement capabilities of our approach.

We make three major contributions. 
First, we introduce a novel representation, D$^3$Fields, that is \textbf{3D}, \textbf{semantic}, and \textbf{dynamic}.
Second, we present a novel and flexible goal specification method using 2D images that incorporate a wide range of styles, contexts, and instances.
Third, our proposed robotic manipulation framework supports a broad spectrum of zero-shot rearrangement tasks.
\vspace{-5pt}

\section{Related Works}

\vspace{-5pt}

\textbf{Foundation Models for Robotics}. Large Language Models (LLMs) have demonstrated promising reasoning capabilities for language. Robotics researchers have used LLMs to generate plans for manipulation~\cite{brohan2023can, huang2023inner, liang2023code, huang2023voxposer}.
Yet, their perception modules fall short in simultaneously modeling the 3D geometry, semantics, and dynamics of objects.
Meanwhile, visual foundation models, such as SAM~\cite{kirillov2023segany} and DINOv2~\cite{oquab2023dinov2}, have demonstrated impressive zero-shot generalization capabilities across various vision tasks.
While prior visual models, like Dense Object Nets~\cite{pmlr-v87-florence18a}, can encode similar semantic information on a small-scale dataset, these foundational models show better generalization capabilities on various object categories and scenarios.
However, their focus is primarily on 2D vision tasks. Grounding these models in a dynamic 3D environment remains a challenge.
Recent works showcase how to ground these foundational models in the 3D world and help imitation learning to generalize~\cite{zhu2023learning, gervet2023act3d, xian2023chaineddiffuser,3d_diffuser_actor}.
Still, these works do not emphasize dynamics learning or achieve zero-shot generalization ability.

\textbf{Neural Fields for Robotic Manipulation.} There are various approaches leveraging neural fields as a representation for robotic manipulation~\cite{zhu2021rgb,ichnowski2021dex,simeonov2022neural,driess2022learning, Jiang-RSS-21, weng2023ngdf, 22-driess-NeRF-RL, pmlr-v202-shim23a, yen2022nerfsupervision, tang2023icra:rgbonly, zhou2023nerf, Shafiullah-RSS-23, wi2022virdo++, wi2022virdo}.
Among them, a series of works distilling neural feature fields from visual foundation models are closely related to us~\cite{pmlr-v139-radford21a, caron2021emerging, lerf2023, sharma2023language, shen2023distilled}.
However, they often require dense camera views for a quality field, which is expensive and impractical for real-world scenarios. Also, distilled neural fields need retraining for new scenes, which is time-consuming and inefficient. In contrast, our D$^3$Fields require no training for new scenes and can work with sparse views and dynamic settings.
GNFactor and FeatureNeRF train neural feature fields that can be conditioned on sparse views~\cite{ze2023multitask, ye2023featurenerf}. However, such fields are often trained on a small dataset, making them hard to generalize to novel instances and scenes, whereas our D$^3$Fields offer better generalization capability to new instances.

\vspace{-5pt}

\section{Method}

\begin{figure*}[t]
    \centering
    \includegraphics[width=\linewidth]{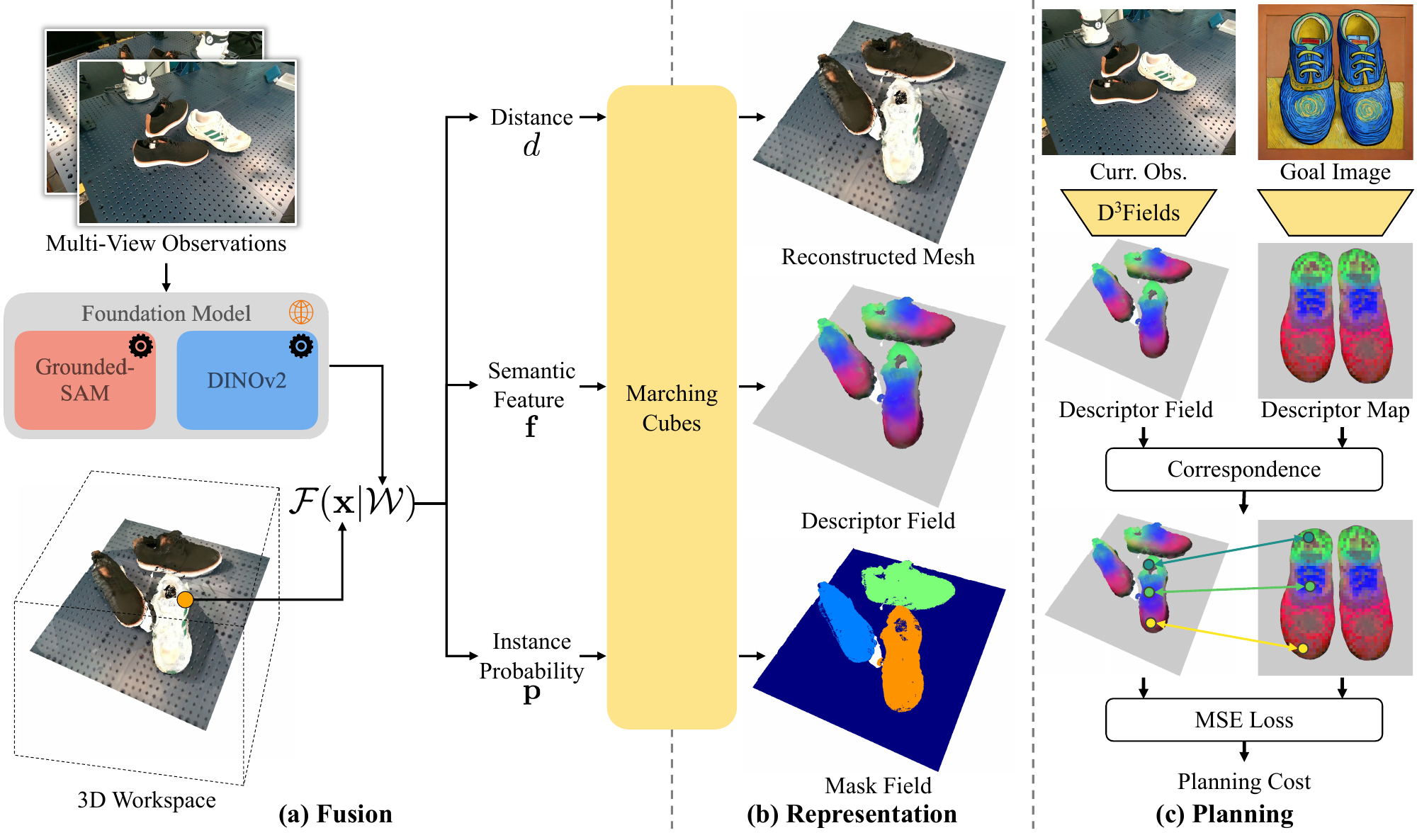}
    \vspace{-20pt}
    \caption{\small
    \textbf{Overview of the Proposed Framework.}
(a) Multi-view RGBD observations are first processed by foundation models to obtain the feature volume $\mathcal{W}$
The implicit function $\mathcal{F}$ takes in arbitrary 3D points and outputs corresponding distance $d$, semantic features $\mathbf{f}$, and instance probability $\mathbf{p}$.
(b) Through marching cubes, we could reconstruct the mesh from the implicit signed distance function.
Since our representation also encodes instances masks and semantic features for evaluated 3D points, we can construct meshes for the mask field and descriptor field as well.
(c) Given a 2D goal image, we use foundation models to extract the descriptor map. Then we correspond 3D features to 2D features and define the planning cost based on the correspondence.
    }
    \label{fig:fusion}
    \vspace{-15pt}
\end{figure*}

\vspace{-5pt}
\subsection{Problem Formulation}
\vspace{-5pt}
\label{sec:problem}

We formulate our problem as a zero-shot rearrangement problem given a 2D goal image $\mathcal{I}$ and RGBD images from multiple fixed viewpoints.
We denote the workspace scene representation as $\bm{s}_{\text{goal}}$.
Our goal is to find an optimal action sequence $\{a^t\}$ to minimize the task objective:
\begin{equation}
\begin{aligned}
\min_{\{a_t\}} \quad &c(\bm{s}^T, \bm{s}_{\text{goal}}), \\
\textrm{s.t.} \quad &\bm{s}^t = g(\bm{o}^t), 
&\bm{s}^{t+1} = f(\bm{s}^{t}, a^t),
\label{eq:obj_fn}
\end{aligned}
\end{equation}
where $c(\cdot,\cdot)$ is the cost function measuring the distance between the terminal representation $\bm{s}^T$ and the goal representation $\bm{s}_{\text{goal}}$. Representation extraction function $g(\cdot)$ takes in the current multi-view RGBD observations $\bm{o}^t$ and outputs the current representation $\bm{s}^t$. $f(\cdot,\cdot)$ is the dynamics function that predicts the future representation $\bm{s}^{t+1}$, conditioned on the current representation $\bm{s}^t$ and action $a^t$. The optimization aims to find the action sequence $\{a_t\}$ that minimizes the cost function $c(\bm{s}^T, \bm{s}_{\text{goal}})$.

Section~\ref{sec:repr} describes how to construct representation extraction function $g(\cdot)$.
Section~\ref{sec:dyn} elaborates on the dynamics function $f(\cdot,\cdot)$.
In Section~\ref{sec:plan}, we show how to define the cost function $c(\cdot,\cdot)$.

\vspace{-5pt}

\subsection{D\texorpdfstring{$^3$}{}Fields Representation}
\label{sec:repr}

\vspace{-5pt}

We assume that we can access multiple RGBD cameras with fixed viewpoints to construct D$^3$Fields.
Multi-view RGBD observations are first fed into visual foundational models. Then we obtain 2D feature volumes $\mathcal{W}$.
D$^3$Fields are implicit functions $\mathcal{F}(\cdot|\mathcal{W})$ defined as follows:
\begin{equation}
\begin{aligned}
    (d, \mathbf{f}, \mathbf{p}) = \mathcal{F}(\mathbf{x}|\mathcal{W}),
\end{aligned}
\end{equation}
where $\mathbf{x}$ can be an arbitrary 3D coordinate in the world frame, and $(d, \mathbf{f}, \mathbf{p})$ corresponds to the signed distance $d \in \mathbb{R}$, the semantic descriptor $\mathbf{f} \in \mathbb{R}^{N}$, and the instance probability distribution $\mathbf{p} \in \mathbb{R}^{M}$ of $M$ instances. $M$ could be different across scenarios.

\begin{wrapfigure}{r}{7cm}
    \vspace{-10pt}
    \centering
    \includegraphics[width=.99\linewidth]{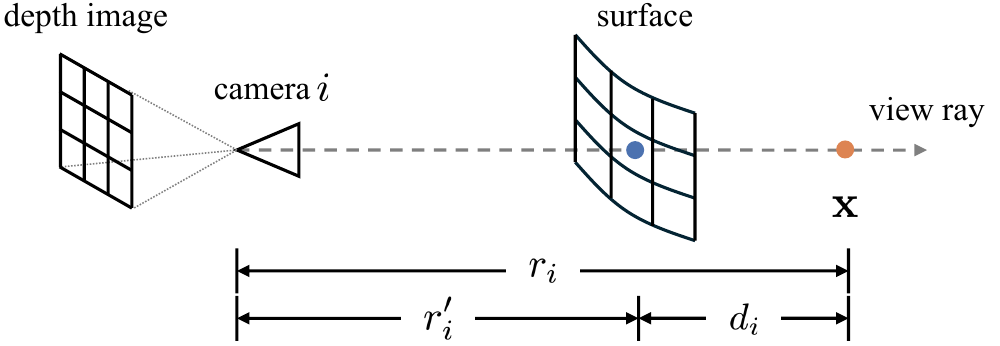}
    \vspace{-3pt}
    \caption{\small
    \textbf{Notation Illustration.}
    $r_i$ is the distance between a 3D point $\mathbf{x}$ and camera $i$, and $r'_i$ is the interpolated depth from the depth image.
    }
    \label{fig:method_explanation}
    \vspace{-10pt}
\end{wrapfigure}

As an overview, our pipeline first projects $\mathbf{x}$ into the image space of each camera.
Then, we can obtain the truncated depth difference between projected depth and real depth reading using Equation~\ref{eq:trunc_depth}.
Afterwards, we assign weights to each viewpoint using Equation~\ref{eq:weights} and interpolate the semantic features and instance masks for each camera using Equation~\ref{eq:single_feat}.
Finally, we fuse features from all viewpoints to obtain the final descriptor using Equation~\ref{eq:fuse}.

More concretely, we map an arbitrary 3D point $\mathbf{x}$ to the $i$th viewpoint's image space.
We denote the projected pixel as $\mathbf{u}_i$ and the distance from $\mathbf{x}$ to the $i$th viewpoint as $r_i$ (Figure~\ref{fig:method_explanation}).
By interpolating the $i$th viewpoint's depth image $\mathcal{R}_i$, we compute the corresponding depth reading from the depth image as $r'_i = \mathcal{R}_i[\mathbf{u}_i]$.
Then we can compute the truncated depth difference as
\begin{equation}
\label{eq:trunc_depth}
\begin{aligned}
    d_i = r_i - r'_i,
    &\quad
    d_{i}' = \textrm{max}(\textrm{min}(d_i, \mu), -\mu),\\
\end{aligned}
\end{equation}
where $\mu$ specifies the truncation threshold for the Truncated Signed Distance Function (TSDF).
Given the truncated depth difference, we compute weights $v_i$ and $w_i$ for each viewpoint as 
\begin{equation}
\label{eq:weights}
\begin{aligned}
    v_i = \mathds{1}_{d_i < \mu},\quad
    w_i = \exp{\left(\frac{\min\left( \mu - |d_i|, 0\right)}{\mu}\right)}.
\end{aligned}
\end{equation}
Here is the explanation and design justification for each term.
\vspace{-8pt}
\begin{itemize}[leftmargin=0.3in]
    \item[$v_i$:] It represents the visibility of $\mathbf{x}$ in camera $i$.
    $\mathds{1}_{d_i < \mu}$ is the indicator function, which equals to 1 when $d_i < \mu$ and equals to 0 otherwise.
    When $d_i = r_i - r'_i \ge \mu$,
    $\mathbf{x}$ is behind the surface, which means $\mathbf{x}$ is not visible in camera $i$ and $v_i=0$.
    \vspace{-5pt}
    \item[$w_i$:] It is the weight for the $i$th viewpoint.
    Since we only have a confident estimation when $\mathbf{x}$ is close to the surface, $w_i$ will decay as $|d_i|$ increases.
    For $\mathbf{x}$ that is far away, $w_i$ degrades to $0$.
\end{itemize}
\vspace{-8pt}
Then we extract the semantic feature $\mathbf{f}_i$ and instance mask $\mathbf{p}_i$ in each viewpoint using
\begin{equation}
\label{eq:single_feat}
\begin{aligned}
    \mathbf{f}_i = \mathcal{W}_i^{\mathbf{f}}[\mathbf{u}_i], &\quad
    \mathbf{p}_i = \mathcal{W}_i^{\mathbf{p}}[\mathbf{u}_i],
\end{aligned}
\end{equation}
where DINOv2~\cite{oquab2023dinov2} extracts the semantic feature volume $\mathcal{W}_i^{\mathbf{f}}\in \mathbb{R}^{H\times W \times N}$ from RGB image. $\mathcal{W}_i^{\mathbf{p}}\in \mathbb{R}^{H\times W \times M}$ is the instance mask volume using Grounded-SAM~\cite{liu2023grounding, kirillov2023segany}. Note that $\bm{p}_i$ is a one-hot vector and already associated to ensure consistent instance indexing across different views.
Finally, we fuse the semantic features and instance masks from all $K$ viewpoints using
\begin{equation}
\begin{aligned}
    d = \frac{\sum_{i=1}^{K} v_i d_{i}'}{\delta + \sum_{i=1}^{K} v_i}, \quad
    \mathbf{f} = \frac{\sum_{i=1}^{K} v_i w_i \mathbf{f}_i}{\delta + \sum_{i=1}^{K} v_i}, \quad
    \mathbf{p} = \frac{\sum_{i=1}^{K} v_i w_i \mathbf{p}_i}{\delta + \sum_{i=1}^{K} v_i},
\end{aligned}
\label{eq:fuse}
\end{equation}
where $\delta$ is a small number to avoid numeric issues. Since the process of projection, interpolation, and fusion is differentiable, $\mathcal{F}(\cdot|\mathcal{W})$ is differentiable when $\mathbf{x}$ is within the truncation threshold.

\subsection{Keypoints Tracking and Dynamics Learning}
\label{sec:dyn}

\vspace{-5pt}

This section will present how to use the dynamic implicit 3D descriptor field $\mathcal{F}(\cdot|\mathcal{W})$ to track keypoints and train dynamics.
Without loss of generality, consider the tracking of a single object instance $\bm{s}^{t}\in \mathbb{R}^{3\times n_{\bm{s}}}$.
For clarity, we denote $\mathbf{f}$ and $d$ from $\mathcal{F}(\cdot|\mathcal{W})$ as $\mathcal{F}_{\mathbf{f}}(\cdot|\mathcal{W})$ and $\mathcal{F}_{d}(\cdot|\mathcal{W})$.
We initialize the tracked keypoints $\bm{s}^0$ by sampling points close to the surface of the desired instance.
To track keypoints $\bm{s}^0$, we formulate the tracking problem as an optimization problem:
\begin{equation}
\begin{aligned}
    \min_{\bm{s}^{t+1}} \quad & ||\mathcal{F}_{\mathbf{f}}(\bm{s}^{t+1}|\mathcal{W}) - \mathcal{F}_{\mathbf{f}} (\bm{s}^0|\mathcal{W})||_2.
\end{aligned}
\end{equation}
As $\mathcal{F}(\cdot|\mathcal{W})$ is differentiable, we could use a gradient-based optimizer. This method could be naturally extended to multiple-instance scenarios.
We found that relying solely on features for tracking can be unstable. Therefore, if we know that the tracked object is rigid, we can apply additional rigid constraints and distance regularization for more stable tracking.

Keypoint tracking enables dynamics model training on real data. We instantiate the dynamics model $f(\cdot,\cdot)$ as graph neural networks (GNNs). We follow~\cite{li2020visual} to predict object dynamics. Please refer to~\cite{li2018learning,li2020visual} for more details on how to train the graph-based neural dynamics model. The trained dynamics model will be used for trajectory optimization in Section~\ref{sec:plan}.

\vspace{-5pt}

\subsection{Zero-Shot Generalizable Robotic Rearrangement}
\label{sec:plan}

\vspace{-5pt}

In this section, we will describe how to define the planning cost for our zero-shot rearrangement framework.
As shown in Figure~\ref{fig:fusion} (c), we first find the correspondence between the descriptor fields and goal image using Equation~\ref{eq:corr}.
Then we define the cost function $c(\cdot,\cdot)$ in Equation~\ref{eq:cost} to measure the distance between the current state and the goal state.
Finally, we optimize the action sequence $\{a_t\}$ to minimize the cost function as described in Section~\ref{sec:problem}.

As described in Section \ref{sec:repr}, we initially sample points $\bm{s}^0\in \mathbb{R}^{3\times n_{\bm{s}}}$ and obtain the associated features $\mathbf{f}^0\in \mathbb{R}^{f\times n_{\bm{s}}}$ from the descriptor fields.
We correspond $\bm{s}^0$ to the goal image $\mathcal{I}_{\text{goal}}$ to define 2D goal points $\bm{s}_{\text{goal}} \in \mathbb{R}^{2 \times n_{\bm{s}}}$.
Firstly, we compute the feature distance $\alpha_{ij}$ between $i$th pixel $\mathbf{u}_i$ of $\mathcal{I}_{\text{goal}}$ and $j$th sampled point of $\bm{s}^0$.
Then we normalize $\alpha_{ij}$ using the softmax over the whole image and obtain the weight $\beta_{ij}$.
Lastly, we find the 2D point $\bm{s}_{\text{goal}, j}$ corresponding to the $j$th 3D point using weighted sum.
The computation process is summarized in the following:
\begin{equation}
\begin{aligned}
\label{eq:corr}
    \alpha_{ij} = ||\mathcal{W}^{\mathbf{f}}_{\text{goal}}[\mathbf{u}_i] - \mathbf{f}_j^0||_2, \quad
    \beta_{ij} =\frac{\exp{\left(-s\alpha_{ij}\right)}}{\sum_{i=1}^{H\times W} \exp{\left(-s\alpha_{ij}\right)} }, \quad
    \bm{s}_{\text{goal}, j} = \sum_{i=1}^{H\times W}\beta_{ij}\mathbf{u}_i,
\end{aligned}
\end{equation}
where $\mathcal{W}^{\mathbf{f}}_{\text{goal}}$ is the feature volume extracted from $\mathcal{I}_{\text{goal}}$ using DINOv2, and $s$ is the hyperparameter to determine whether the heatmap $\beta_{ij}$ is more smooth or concentrating. Although Equation.~\ref{eq:corr} only shows a single instance case, it could be naturally extended to multiple instances by using instance mask information.

\begin{wrapfigure}{r}{6cm}
    \vspace{-24pt}
    \centering
    \includegraphics[width=.99\linewidth]{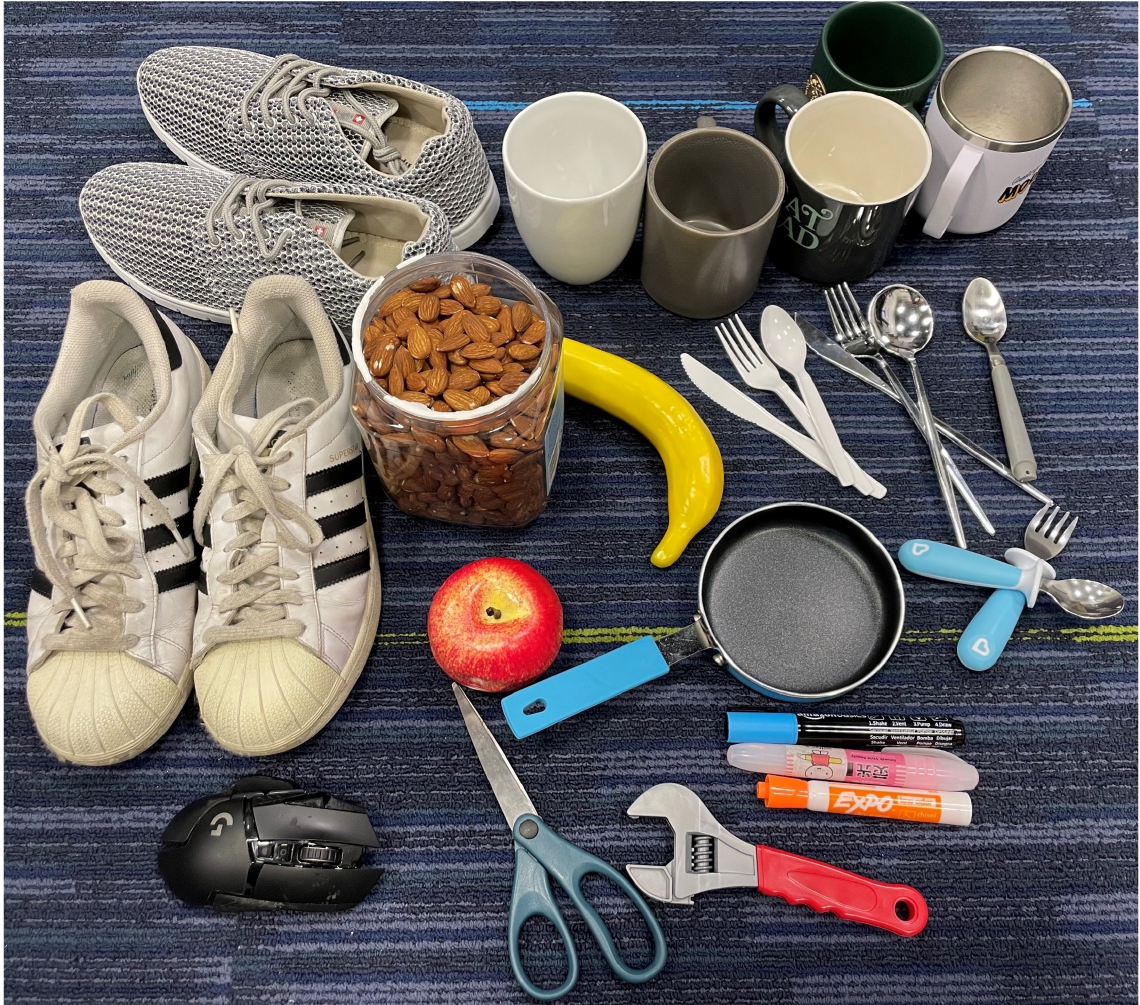}
    \vspace{-17pt}
    \caption{\small
    \textbf{Object Set in Our Experiments.}
    This figure shows diverse objects used in our experiments, expanding over 10 object types.
    }
    \label{fig:objs}
\end{wrapfigure}

Note that
$\bm{s}_\text{goal}$ is in the image space, and the current state representation, $\bm{s}^t$ is in 3D space.
To reconcile this discrepancy, we introduce a \textit{virtual} reference camera.
We project $\bm{s}^t$ into the reference image and obtain its 2D positions $\bm{s}^t_{\text{2D}}$.
Consequently, we define the task cost function as follows:%
\begin{equation}
\label{eq:cost}
\begin{aligned}
    c(\bm{s}^t, \bm{s}_{\text{goal}}) = ||\bm{s}^t_{\text{2D}} - \bm{s}_{\text{goal}}||_2^2.
\end{aligned}
\end{equation}
Given the planning cost, we could use an MPC framework to derive the action sequence $\{a_t\}$ to minimize the cost function.
Specifically, we use MPPI to optimize the action sequence~\cite{williams2017information}.
At each time step, we sample a set of action sequences and evaluate the cost function.
Then we update the action sequence using the cost function and repeat the process until convergence.
This process will be repeated for each time step until the task is completed.
More details regarding the method are included in the supplementary material.

\vspace{-5pt}
\section{Experiments}
\vspace{-10pt}

We design and organize our experiments to answer three key questions about our method:
(1)~How efficient and effective our D$^3$Fields is compared to existing neural representations?
(2)~What kind of manipulation tasks can be enabled by our framework, and what type of generalization can it achieve?
(3)~Why can our D$^3$Fields enable these tasks and be generalizable?

\vspace{-8pt}
\subsection{Experiment Setup}
\label{sec:setup}

\vspace{-7pt}

In the real world, we use four RGBD cameras positioned at the corners of the workspace and employ the Kinova\textsuperscript{\textregistered} Gen3 arm for action execution.
In the simulation, we use OmniGibson and the Fetch robot for mobile manipulation tasks~\cite{li2022behavior}. Our evaluations span various tasks, including organizing shoes, collecting debris, tidying the table, and so on.
More details are in the supplementary material.

\vspace{-8pt}

\subsection{D\texorpdfstring{$^3$}{}Fields Efficiency and Effectiveness}
\label{sec:comp}

\vspace{-7pt}

\begin{figure}[!t]
    \centering
    \includegraphics[width=\linewidth]{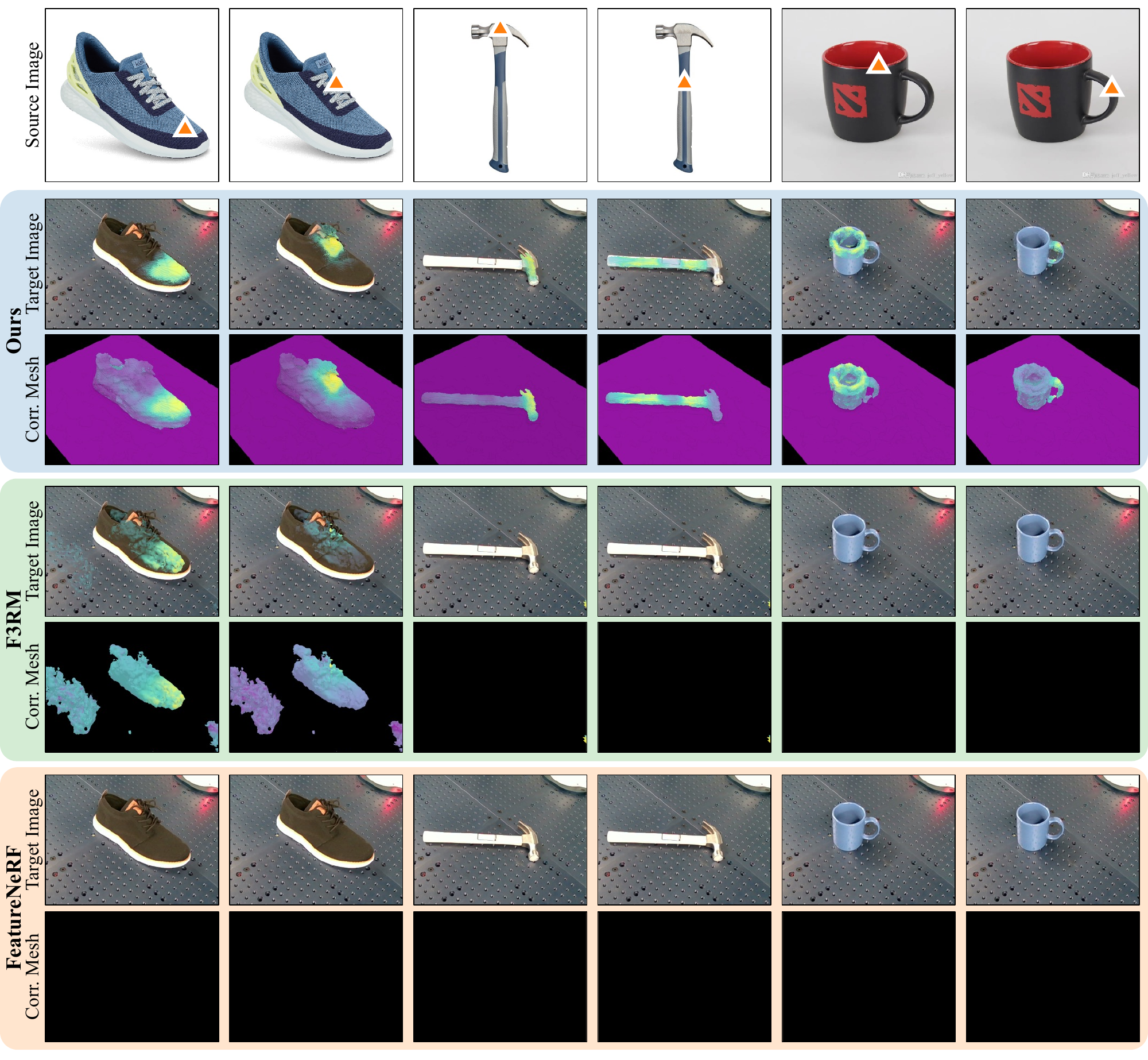}
    \vspace{-20pt}
    \caption{\small
    \textbf{Correspondence Qualitative Comparison.}
    We select the pixel from the source image, obtain the associated DINOv2 feature, and visualize the correspondence heatmap on the reconstructed mesh.
    (a) Our representation reconstructs clear mesh and corresponds from the source image to semantically similar 3D areas.
    (b) F3RM~\cite{shen2023distilled} could construct a reasonable mesh in the shoe scene and establish rough correspondences, but fails in other scenes.
    (c) Only trained on a small dataset, FeatureNeRF~\cite{ye2023featurenerf} fails to generalize to novel scenes. The reconstructed meshes are out of camera view, and the correspondence quality is poor.
    }
    \label{fig:corr_quali_comp}
    \vspace{-15pt}
\end{figure}

In this section, we compare our D$^3$Fields with two baselines, Distilled Feature Fields (F3RM)~\cite{shen2023distilled} and FeatureNeRF~\cite{ye2023featurenerf}.
For F3RM, we use four camera views as inputs and the DINOv2 features as the supervision and stop the distillation process at 2,000 steps, which is defaulted by the authors~\cite{shen2023distilled}.
We first compare our D$^3$Fields with F3RM and FeatureNeRF in terms of the correspondence accuracy, as shown in Figure~\ref{fig:corr_quali_comp}.
We reconstruct the mesh from our 3D implicit representation using marching cubes.
We then select DINOv2 features from the source image and visualize the corresponding heatmap on the reconstructed feature mesh.
We could see that our D$^3$Fields could reconstruct the mesh with high quality and highlight semantically similar areas across different instances and contexts, while F3RM fails to reconstruct the mesh accurately, which leads to an inaccurate correspondence heatmap.
Since FeatureNeRF is only pre-trained on a small dataset, it fails to construct an accurate mesh of the scene and find accurate correspondence in novel scenes, demonstrating our representation's effectiveness in reconstructing meshes and encoding semantic features. Quantitative comparisons are included in the supplementary material.

We also evaluate the time efficiency of constructing the implicit representation given four RGBD views across four scenes on the machine with A6000 GPU. Our method takes $0.166 \pm 0.002$ seconds, which is significantly more efficient than F3RM, which takes $88.379 \pm 5.306$ seconds.

\vspace{-11pt}

\subsection{Zero-Shot Generalizable Rearrangement}
\label{sec:tasks}

\vspace{-9pt}

\begin{figure*}[t]
    \centering
    \includegraphics[width=.99\linewidth]{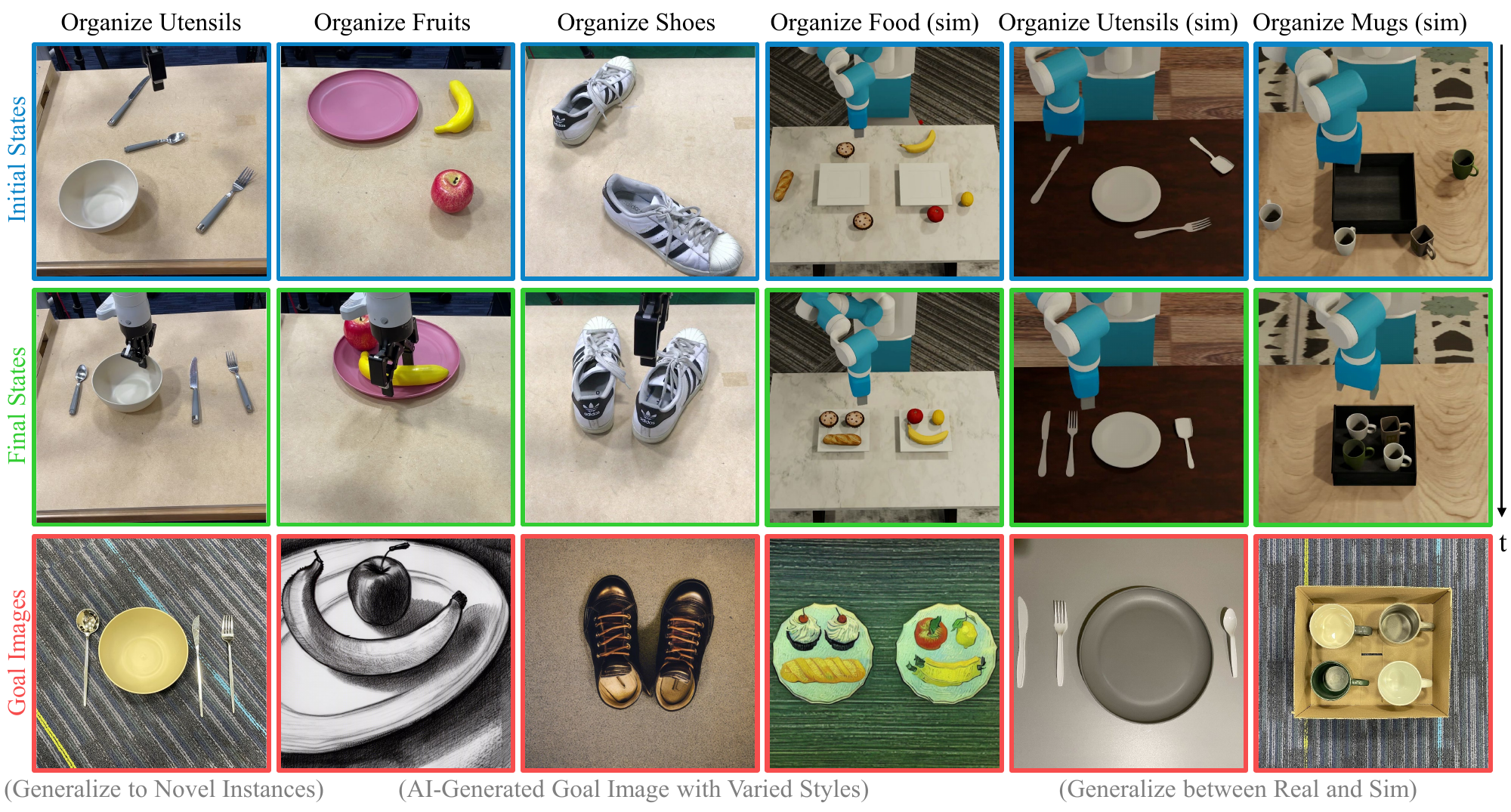}
    \vspace{-8pt}
    \caption{\small
    \textbf{Qualitative Results.}
    We qualitatively evaluate our proposed framework on rearrangement tasks, both in the real world and in simulation, encompassing tasks such as organizing utensils, fruits, shoes, food, and mugs.
    The figure highlights that our representation can generalize across varied instances, styles, and contexts.
    For instance, in the example of organizing fruit, the goal image, unlike the workspace, is styled as a sketch drawing.
    Because our representation encodes semantic features, the banana in the workspace can correspond with the banana in the goal image, which allows the task to be finished.
    This wide range of tasks showcases the generalization capabilities of our framework.
    }
    \label{fig:quali}
    \vspace{-20pt}
\end{figure*}

We conduct a qualitative evaluation of D$^3$Fields in common robotic rearrangement tasks in a zero-shot manner, with partial results displayed in Figure~\ref{fig:teaser} and Figure~\ref{fig:quali}. We observed several key capabilities of our framework, which are as follows:

\vspace{-6pt}

\textbf{Generalization to AI-Generated Goal Images.} In Figure~\ref{fig:teaser}, the goal image is rendered in a Van Gogh style, which is distinctly different from those in the workspace. Since D3Fields encode semantic information, capturing shoes with varied appearances under similar descriptors, our framework can manipulate shoes based on AI-generated goal images.

\vspace{-6pt}

\textbf{Compositional Goal Images.} 
In the office desk organization task in Figure~\ref{fig:teaser}, the robot first arranges the mouse and pen to match the goal image. Then, the robot repositions the mug from the top of a box to the mug pad, guided by a separate goal image of the upright mug. This example illustrates that our system can accomplish tasks using compositional goal images.

\vspace{-6pt}

\textbf{Generalization across Instances and Materials.} 
Figure~\ref{fig:objs} and Figure~\ref{fig:teaser} also show our framework's ability to generalize across various instances and materials. For example, the debris collection in Figure~\ref{fig:teaser} shows our framework's ability to handle granular objects. Figure~\ref{fig:quali} further shows our framework’s instance-level generalization, where the goal instances distinct from ones in the workspace.

\vspace{-6pt}

\textbf{Generalization across Simulation and Real World.} We evaluated our framework in the simulator, as shown in the utensil organization and mug organization examples in Figure~\ref{fig:quali}. Given goal images from the real world, our framework can also manipulate objects to the goal configurations. This underscores our framework's generalization capabilities between simulation and the real world.

\vspace{-10pt}

\subsection{D\texorpdfstring{$^3$}{}Fields Analysis}
\label{sec:analysis}

\vspace{-8pt}

\begin{figure*}[t]
    \centering
    \includegraphics[width=\linewidth]{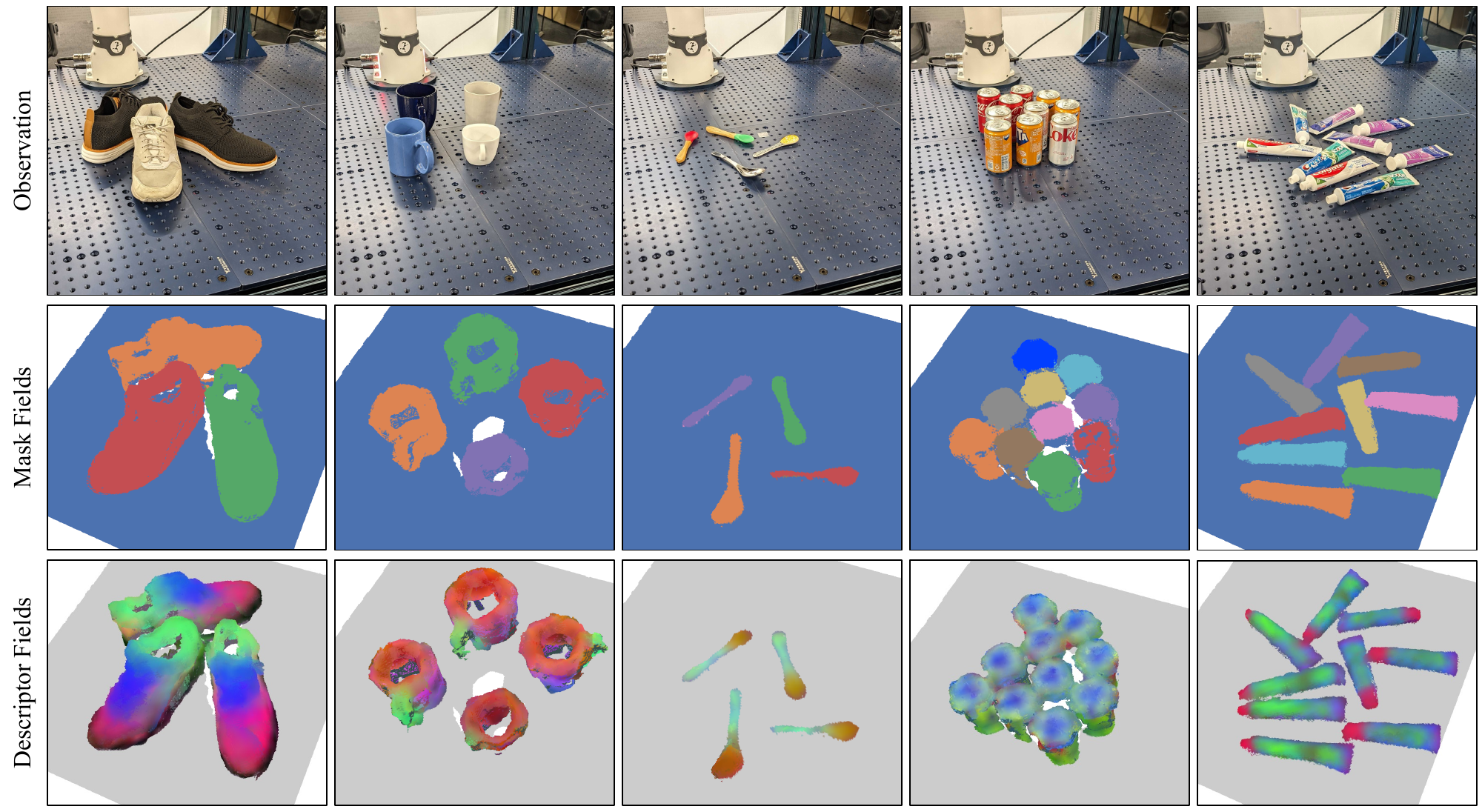}
    \vspace{-15pt}
    \caption{\small
    \textbf{Representation Visualizations.}
    To analyze our representation,
    we visualize the representation across different object categories.
Mask fields color objects based on their instance masks, showing effective distinguishing among different instances.
Descriptor fields color 3D points through PCA.
The consistent color patterns across instances show that our representation could encode consistent semantic features over different instances.
We also demonstrate our representation's robustness to clustered scenes in the right two columns.
    }
    \label{fig:pca}
    \vspace{-15pt}
\end{figure*}

To obtain a more thorough understanding of our D$^3$Fields, we first extract the mesh using the marching cube algorithm. We evaluate vertices from the extracted mesh using our D$^3$Fields and obtain the associated segmentation information and semantic features, as visualized in Figure~\ref{fig:pca}.
Mask fields in Figure~\ref{fig:pca} show a distinct 3D instance segmentation in different scenarios, even clustering scenes like cans and toothpaste.
The 3D instance segmentation enables the downstream planning module to distinguish and manipulate multiple instances, as shown in the mug organization tasks.

\vspace{-3pt}

Additionally, we visualize the semantic features by mapping them to RGB space using PCA. We observe that our semantic fields show consistent color patterns across different instances.
In the provided shoe example, even though various shoes have distinct appearances and poses, they exhibit similar color patterns: shoe heels are represented in green, and shoe toes in red. We observed similar patterns for other object categories such as mugs and forks.
The consistent semantic features across various instances help our manipulation framework to achieve category-level generalization. When encountering novel instances, our D$^3$Fields can still establish the correspondence between the workspace and the goal image using semantic features.
In addition, we analyze in the supplementary to show that D$^3$Fields have better 3D consistency compared to simple point cloud stitching.

\vspace{-5pt}

\vspace{-5pt}

\section{Conclusion}

\vspace{-10pt}

In this work, we introduce D$^3$Fields, which implicitly encode 3D semantic features and 3D instance masks, and
model the underlying dynamics.
Our emphasis is on zero-shot generalizable rearrangement tasks specified by 2D goal images of varying styles, contexts, and instances.
Our framework excels in executing a diverse array of robotic rearrangement tasks in both simulated and real-world scenarios.
Its performance greatly surpasses baseline methods such
as FeatureNeRF and F3RM in terms of efficiency and effectiveness.

\vspace{-10pt}
\paragraph{Limitation.}

Our method lifts 2D visual foundation models to the 3D world, enabling a range of zero-shot rearrangement manipulation tasks. However, we both benefit from and are limited by their capabilities. For example, the semantic feature field is not fine-grained enough to distinguish between the right and left sides of shoes.
Visual foundation models with more fine-grained semantic features are needed. In addition, rearrangement tasks could fail when they need to follow a specific manipulation order to avoid collision with other objects, e.g. a crowded scene. In the future, incorporating a task planner to handle clustered scenes could be a valuable direction.

\acknowledgments{This work is partially supported by Sony Group Corporation, the Stanford Institute for Human-Centered AI (HAI), and Google. The opinions and conclusions expressed in this article are solely those of the authors and do not reflect those of the sponsoring agencies.}

\bibliography{references}

\end{document}